\documentclass{article}
\usepackage{ijcai21}

\usepackage{times}

\usepackage{soul}
\usepackage{url}
\usepackage[hidelinks]{hyperref}
\usepackage[utf8]{inputenc}
\usepackage[small]{caption}
\usepackage{graphicx}
\usepackage{amsmath}
\usepackage{booktabs}
\usepackage[ruled,vlined]{algorithm2e}
\usepackage{tablefootnote}
\usepackage{subfigure}
\usepackage{booktabs}
\usepackage{multirow}
\usepackage{footnote}
\usepackage{threeparttable}
\usepackage{tablefootnote}
\title{Auto-weighted low-rank representation for clustering}

\author{Zhiqiang Fu$^{1,2}$\and
Yao Zhao$^{1,2}$\footnote{Contact Author}\and
Dongxia Chang$^{1,2}$\and
Xingxing Zhang$^3$ \And Yiming Wang$^{1,2}$\\
\affiliations
$^1$Institute of Information Science, Beijing Jiaotong University, Beijing\\
$^2$Beijing Key Laboratory of Advanced Information Science and Network Technology, Beijing\\
$^3$Department of Computer Science and Technology, Tsinghua University\\
\emails
\{zhiqiangfu, yzhao, dxchang\}@bjtu.edu.cn,xxzhang2020@mail.tsinghua.edu.cn, wangym@bjtu.edu.cn
}

\begin{document}

\maketitle

\begin{abstract}
In this paper, a novel unsupervised low-rank representation model, i.e., Auto-weighted Low-Rank Representation (ALRR), is proposed to construct a more favorable similarity graph (SG) for clustering. In particular, ALRR enhances the discriminability of SG by capturing the multi-subspace structure and extracting the salient features simultaneously. Specifically, an auto-weighted penalty is introduced to learn a similarity graph by highlighting the effective features, and meanwhile, overshadowing the disturbed features. Consequently, ALRR obtains a similarity graph that can preserve the intrinsic geometrical structures within the data by enforcing a smaller similarity on two dissimilar samples. Moreover, we employ a block-diagonal regularizer to guarantee the learned graph contains $k$ diagonal blocks. This can facilitate a more discriminative representation learning for clustering tasks. Extensive experimental results on synthetic and real databases demonstrate the superiority of ALRR over other state-of-the-art methods with a margin of 1.8\%$\sim$10.8\%.

\end{abstract}

\section{Introduction}
Clustering is a key technique of data mining, which targets at grouping the given database automatically without any label information. Many clustering methods have been proposed in past decades, e.g.,\cite{ShahK17,ZhangXX2017,ZhangXX2018,ZhangXX2019,WenZ0ZFL21}. Due to the good performance and strong theory basis, graph-based clustering, which is a critical branch of clustering methods, has become an attractive research area. In general, most graph-based clustering methods can be summarized as two steps. First, a similarity graph (SG) should be constructed to depict the pairwise relations among samples. Then, this graph is divided into $k$ sub-graphs according to a strategy, where $k$ is the number of clusters. Therefore, the performance of the graph-based clustering strongly depends on the quality of the SG. However, constructing a discriminative SG is difficult for high-dimensional data because many metric methods become worse as the dimension increases.

Recently, self-representation theories have pointed out that a given high-dimensional samples can be regarded as sampled from $k$ independent low-dimensional subspaces, and each sample can be linearly represented by the other samples in the same subspace \cite{LiuXY12}. Based on this assumption, many self-representation methods have been proposed to construct the SG by learning the subspace structure of samples, e.g., low-rank representation (LRR) \cite{liu2012robust} and FLLRR \cite{SongW18}. 

LRR is an important branch of self-representation models and has a strong theory basis. It tries to capture the subspace structure with a global low-rank constraint. However, LRR can learn the global structure but ignore the local geometry information \cite{Fei2017Low}. To overcome this problem, a non-negative low-rank learning method \cite{he2011nonnegative} is proposed to capture the local structure by introducing a $L_1$-norm constraint which can ensure the representation using the nearby samples as much as possible. Futhermore, this method also enhance the physical meaning of SG by ensuring that the SG is nonnegative. Motivated by manifold learning \cite{belkin2008towards}, Laplacian regularized LRR \cite{yin2015laplacian} is proposed to learn more local structure by adopting a Laplacian regularization which ensures that the samples similar in the original space are also similar in representation space. Moreover, RSEC \cite{tao2019robust} is proposed to improve the clustering results of by adopting a clustering constraint to enhance the discriminability of the learned SG.

All the LRR methods mentioned above usually base on an assumption that the importance of each feature is equal. However, the importance of features is different in real applications \cite{WangHHNCM20}. Moreover, in the clustering task, there has no prior information to previously set reasonable weights to features. To alleviate these problems, we propose a auto-weighted low-rank representation (ALRR) method, and our main contributions are summarized as follows. 
\begin{enumerate}
\setlength{\itemsep}{0pt}
\setlength{\parsep}{0pt}
\setlength{\parskip}{0pt}
    \item We develop ALRR: an Auto-weighted Low-Rank Representation, to improve the discrimination of SG for clustering tasks.
    \item To learn a similarity graph that can preserve the intrinsic geometrical structures within the data, an auto-weighted penalty is introduced by highlighting the effective features and overshadowing the disturbed features.
    \item We employ a block-diagonal regularizer to guarantee the learned graph contains $k$ diagonal blocks, thus facilating a more discriminative representation learning.
    \item An iterative algorithm is developed to solve ALRR. The advantages of our method have been proved by experimental results on synthetic and real databases.
\end{enumerate}
\vspace{-1em}

\section{Notations and Preliminary}
\subsection{Notations}
In this paper, $x_i$ and $x^j$ are the $i$th column and $i$th row of the $X$, respectively. $x_{i,j}$ denotes the element which is on the $i$th row and $j$th column of $X$. $\|x_i\|_2$ is the $l_2$-norm of the vector $x_i$. $X^T$ is the transpose of $X$. $X^{-1}$ is the inverse of $X$. $\text{rank}(X)$ is the rank of $X$. $\text{tr}(X)$ is the trace of $X$. $\|X\|_1$, $\|X\|_F$ and $\|X\|_*$ denote $L_1$-norm, Frobenius norm and nuclear norm of $X$ respectively. $\mathbf{1}$ is the vector in which elements are $1$. $I$ is the identity matrix.
\subsection{Block Diagonal Constraint}
Suppose a ideal database $X_0$ which is strictly sampled from $k$ independent subspaces without any noise. LRR can learn a $k$ block-diagonal SG as $Z_0=V_0V_0^T$, where $V_0$ is obtained by singular value decomposition $X_0=U_0\Sigma_0 V_0^T$. Since there is always noise in the data, a $k$-block diagonal regularizer is proposed to ensure that the matrix contains $k$ diagonal blocks \cite{LuFLMY19}.

\noindent \textbf{Definition 1} ($k$-Block Diagonal Regularizer) For a given SG matrix $B\in R^{n\times n}$, $k$-block diagonal regularizer is defined as 
\begin{equation}
\|B\|_{\boxed{k}}=\sum_{i=n-k+1}^{n} \lambda_{i}\left(L_{B}\right)\label{block}
\end{equation}
where $L_B$ denotes the Laplacian matrix of $B$ and $\lambda_i(L_B)$ is the $i$-th smallest eigenvalue of $L_B$.

\section{The Proposed Method}
\subsection{ALRR: The Objective Function}
For a given database $X=[x_1,x_2,...,x_n]\in R^{d\times n}$ which contains $n$ $d$-dimensional samples, we suppose it contains $k$ clusters sampled from $k$ independent subspaces. As analyzed before, LRR can explore the global low-rank structure but ignores the geometrical structure and the difference in importance of the features. Motivated by this, an auto-weighted matrix is introduced to learn the importance of different features by assigning different weights to the features adaptively. Based on this weighted features, the auto-weighted penalty is employed to preserve the geometrical structure as
\begin{equation}
\begin{split}
    \min_{A,Z,E}\underbrace{\sum_{i,j}\|Ax_i-Ax_j\|_2^2z_{i,j}}_{\text{Auto-weighted~penalty}} +\lambda_1\|Z\|_*+\lambda_2\|E\|_1,\\\text{s.t.}X=XZ+E,Z\ge0,z_{i,i}=0,Z\mathbf{1}=\mathbf{1},\\A = \text{diag}(a),a\ge0,a\mathbf{1}=1 \label{fina}
\end{split}
\end{equation}
where $a\in R^{d}$ is the auto-weighted vector, $A\in R^{d \times d}$ is the auto-weighted matrix which is a diagonal matrix whose mainly diagonal vector is $a$. $Z$ is the SG and $E$ is the recovering error. $\lambda_1$, $\lambda_2$ are two parameters to balance the effect of the three terms. $Ax_i$ is the auto-weighted features of $x_i$, and the auto-weighted matrix $A$ can enhance the original features by assigning large weights to the useful features and assigning small weights to the useless features. Therefore, the weighted features are more discriminative. Based on the auto-weighted features, the auto-weighted penalty can enlarge learned similarity between two samples if the auto-weighted features of them are similar. Thus, this term can also preserve more geometry structure, leading to a more discriminative SG. $a\ge0$ and $a\mathbf{1}=1$ can ensure that the auto-weighted matrix is nonnegative and avoid the trivial solution, i.e., $A=0$. 

As the database is sampled from $k$ independent subspaces, LRR theory has shown that an ideal SG obtained by LRR is a symmetric $k$ block-diagnoal matrix, but this structure usually be destroyed by the noise \cite{FengLXY14}. Since most graph-based clustering methods require symmetric SGs, the learned SGs are usually handled as $W= (|Z|+|Z|^T)/2$ \cite{YangLWRY20}. Hence, to improve the clustering performance, we further introduce a block diagonal constraint to ensure that $W$ have $k$ diagonal blocks as
\begin{equation}
\begin{split}
    \min_{A,Z,E}\underbrace{\sum_{i,j}\|Ax_i-Ax_j\|_2^2z_{i,j}}_{\text{Auto-weighted~penalty}} +\lambda_1\|Z\|_*+\lambda_2\|E\|_1+\\\underbrace{\lambda_3\|\frac{Z+Z^T}{2}\|_{\boxed{k}}}_{\text{Block~constraint}},\text{s.t.}X=XZ+E,Z\ge0,\\z_{i,i}=0,Z\mathbf{1}=\mathbf{1},A = \text{diag}(a),a\ge0,a\mathbf{1}=1 \label{final}
\end{split}
\end{equation}
Utilizing the block diagonal constraint, the $(Z+Z^T)/{2}$ will contain $k$ blocks, which is more suitable for the graph-based clustering methods.

\subsection{ALRR: Optimization}
In this section, we use the ADMM to solve our model. First, two variables, i.e., $S$ and $U$, are introduced, then the Lagrange function of the problem (\ref{final}) can be obtained as
\begin{equation}
\begin{split}
    \min_{A,Z,E,S,U,C_1,C_2,C_3}\sum_{i,j}\|Ax_i-Ax_j\|_2^2s_{i,j} +\lambda_1\|U\|_*+\\\lambda_2\|E\|_1+\lambda_3\|\frac{S+S^T}{2}\|_k+\frac{\mu}{2}(\|X-XZ-E+\\\frac{C_1}{\mu}\|_F^2+\|Z-S+\frac{C_2}{\mu}\|_F^2+\|Z-U+\frac{C_3}{\mu}\|_F^2) \label{obj}
\end{split}
\end{equation}
where $C_1$, $C_2$ and $C_3$ are Lagrange multipliers, and $\mu>0$ is a non-negative penalty. This problem can be divided into several subproblems as follows.

\textbf{1) Update $Z$ with $A$, $E$, $U$ and $S$ fixed.} $Z$ can be obtained by minimizing the following problem
\begin{equation}
\begin{split}
    \min_{Z}\frac{\mu}{2}(\|XZ-M_1\|_F^2+\|Z-M_2\|_F^2+\|Z-M_3\|_F^2)
\end{split}
\end{equation}
where $M_1 = X-E+\frac{C_1}{\mu}$,$M_2=S-\frac{C_2}{\mu}$ and $M_3 = U-\frac{C_3}{\mu}$. By setting the derivative of this formulation to 0, $Z$ can be calculated by a closed form solution as
\begin{equation}
\begin{split}
    Z = (X^TX+2I)^{-1}(X^TM_1+M_2+M_3)\label{Ze}
\end{split}
\end{equation}

\textbf{2) Update $E$ with $A$, $Z$, $U$ and $S$ fixed.} $E$ can be updated by 
\begin{equation}
    \min_E \lambda_2\|E\|_1+\frac{\mu}{2}\|X-XZ-E+\frac{C_1}{\mu}\|_F^2
\end{equation}
According to \cite{LinLS11}, this problem can solved by 
\begin{equation}
\begin{split}
 E = \Omega_{\frac{\lambda_2}{\mu}}(X-XZ+\frac{C_1}{\mu})
\end{split}\label{Ee}
\end{equation}
where $\Omega$ is the shrinkage operator \cite{liu2010robust}.

\textbf{3) Update $U$ with $A$, $Z$, $E$ and $S$ fixed.} By fixing the other variables, $U$ can be obtained by optimizing the following subproblem as
\begin{equation}
 \min_U \lambda_1\|U\|_* + \frac{\mu}{2}\|Z-U+\frac{C_3}{\mu}\|_F^2
\end{equation}
This problem has a closed form solution as 
\begin{equation}
U=\Theta_{\frac{\lambda_{1}}{\mu}}(Z+\frac{C_{3}}{ \mu})\label{Ue}
\end{equation}
where $\Theta$ is the singular value thresholding (SVT) shrinkage operation \cite{liu2012robust}.

\textbf{4) Update $S$ with $A$, $Z$, $U$ and $E$ fixed.} $S$ can be obtained by solving the following subproblem
\begin{equation}
\begin{split}
\min_S \sum_{i,j}\|Ax_i-Ax_j\|_2^2s_{i,j}+\lambda_3\|\frac{S+S^T}{2}\|_k+\\\frac{\mu}{2}\|Z-S+\frac{C_2}{\mu}\|_F^2,\text{s.t.} S\ge 0, s_{i,i}=0, S\mathbf{1}= \mathbf{1}
\end{split}
\end{equation}
Since the definition of the block diagonal regularizer is shown as Eq.(\ref{block}), then we have 
\begin{equation}
\begin{split}
\min_{Z,Y} \|\frac{S+S^T}{2}\|_k=\min_Z  \langle L_S,Y \rangle,\\\text{s.t.}0 \preceq Y \preceq I,  \text{rank}(Y)=k
\end{split}
\end{equation}
where $L_S$ is the Laplacian matrix of $S$, which is defined as $\text{Diag}((S\mathbf{1}+S^T\mathbf{1})/2)-(S+S^T)/2$. Thus, updating $S$ can be divided into two steps. 

First, fix $S$ and update $Y$ by
\begin{equation}
\min_{Y} \langle L_S,Y \rangle,\\\text{s.t.}0 \preceq Y \preceq I,  \text{rank}(Y)=k\label{Ye}
\end{equation}
where $Y$ can be updated by $Y=FF^T$, and $F\in R^{n\times k}$ consists of $k$ eigenvectors associated with the smallest $k$ eigenvalues of $L_S$.

Second, fix $Y$ and update $S$ by
\begin{equation}
\begin{split}
\min_S \sum_{i,j}\|Ax_i-Ax_j\|_2^2s_{i,j}+\lambda_3\langle L_S,Y\rangle+\frac{\mu}{2}\|Z-\\S+\frac{C_2}{\mu}\|_F^2,\text{s.t.} S\ge 0, s_{i,i}=0, S\mathbf{1}= \mathbf{1}
\end{split}
\end{equation}
This problem is equivalent to
\begin{equation}
\begin{split}
\min_S \text{tr}(D_A^TS)+\frac{\mu}{2}\|Z-S+\frac{C_2}{\mu}+\frac{\lambda_3}{2\mu}(D_Y+\\D_Y^T)\|_F^2,\text{s.t.} S\ge 0, s_{i,i}=0, S\mathbf{1}= \mathbf{1}
\end{split}
\end{equation}
where ${d_A}_{i,j}=\|Ax_i-Ax_j\|_2^2$ and $D_Y=\text{diag}(Y)\mathbf{1}^T-Y$.
We can learn a latent variable $\Bar{S}$ without constraint by
\begin{equation}
\Bar{S} = \frac{2\mu Z+2C_2+\lambda_3(D_Y+D_Y^T)-2D_A}{2\mu}
\end{equation}
then $S$ can be obtained by
\begin{equation}
\begin{split}
 s^i = \text{max}(\sigma^i\hat{\mathbf{1}}_i+\Bar{s}^i,0)
\end{split}\label{Se}
\end{equation}
where $\hat{\mathbf{1}}_i$ is a vector that the $i$-th element is 0 and the other elements are 1. $\sigma$ is the Lagrangian multiplier which is defined as
\begin{equation}
    \sigma^{i}=\left(1+\bar{s}^{i} \mathbf{1}\right) /(n-1)
\end{equation}

\textbf{5) Update $A$ with $E$, $Z$, $U$ and $S$ fixed.} $A$ is the auto-weighted matrix which can be obtained by
\begin{equation}
\begin{split}
    \min_A \sum _{i,j} \|Ax_i-Ax_j\|_2^2s_{i,j},\\\text{s.t.}A=\text{diag}(a),a\ge 0, a\mathbf{1}=1
\end{split}
\end{equation}
$A$ can be obtained directly by
\begin{equation}
a_i = \frac{1}{w_i\sum_{i=1}^d\frac{1}{w_i}}
\label{Ae}
\end{equation}
where $w_i=x_iL_Sx_i^T$.

\textbf{6) Update the other variables.}
\begin{equation}
\begin{split}
&C_1 = C_1+\mu(X-XZ-E),\\\label{C1}
&C_2 = C_2+\mu (Z-S),\\
&C_3 = C_3+\mu (Z-U),\\
&\mu = \min(\mu_{\text{max}},\rho \mu).\\
\end{split}
\end{equation}
where $\mu_{\text{max}}$ and $\rho$ are two constants. For convenience, our algorithm is summarized as Algorithm \ref{A1}.
\vspace{-1em}
\begin{algorithm}[htbp]\small
\caption{Solving ALRR}
\label{A1}
\LinesNumbered 
\KwIn{Data matrix $X$ and parameters $\lambda_1$, $\lambda_2$, $\lambda_3$}
\KwOut{$Z$, $A$, $Y$, $E$}
\SetKwData{a}{x}
\textbf{Initialization:} Initializing $Z$ by constructing the $k$-nearest neighbor graph, $S = Z$, $U=Z$, $E = 0$, $C_1 = 0$, $C_2 = 0$, $C_3 = 0$, $\mu = 0.01$, $\rho = 1.1$, $\mu_{max} = 10^8$\;
\While{not converged}{
    Update $Z$ by Eq.(\ref{Ze})\;\label{step1}
    Update $E$ by Eq.(\ref{Ee})\;\label{step2}
    Update $U$ by Eq.(\ref{Ue})\;\label{step3}
    Update $Y$ by Eq.(\ref{Ye})\;\label{step5}
    Update $S$ by Eq.(\ref{Se})\;\label{step4}
    Update $A$ by Eq.(\ref{Ae})\;\label{step6}
    Update $\mu$, $C_1$, $C_2$, $C_3$ by Eq.(\ref{C1}).
    }
\end{algorithm}
\vspace{-1em}
\subsection{Computational Complexity and Convergence Study}
\begin{figure*}[htbp]
\centering
\subfigure[Original]{\begin{minipage}[t]{0.102\linewidth}
\centering
\includegraphics[width=0.8in]{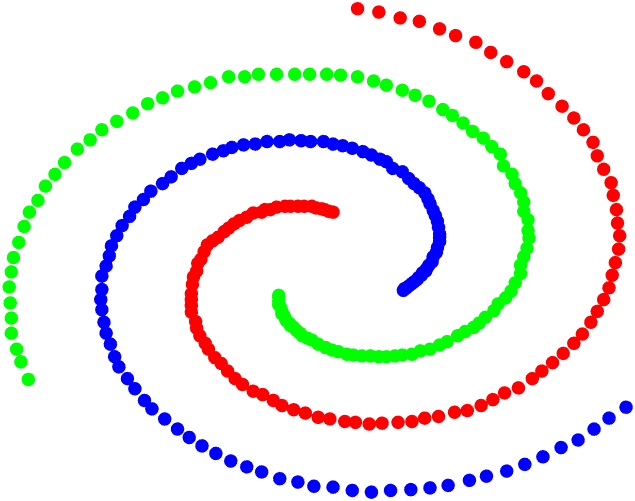}\label{ori}
\end{minipage}}
\quad
\subfigure[LRR]{\begin{minipage}[t]{0.102\linewidth}
\centering
\includegraphics[width=0.8in]{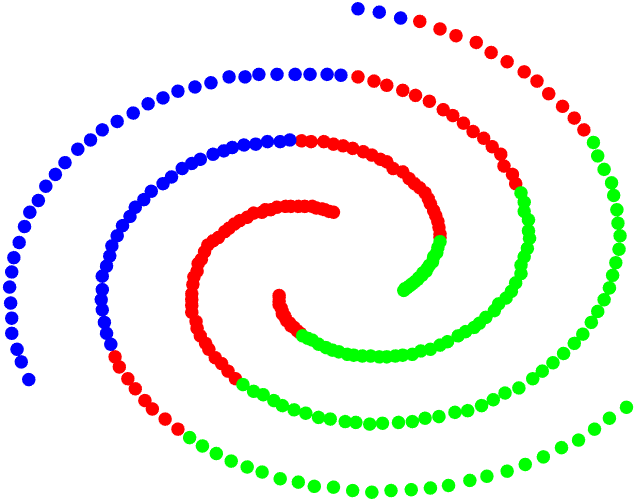}
\end{minipage}}
\quad
\subfigure[NSLLRR]{\begin{minipage}[t]{0.102\linewidth}
\centering
\includegraphics[width=0.8in]{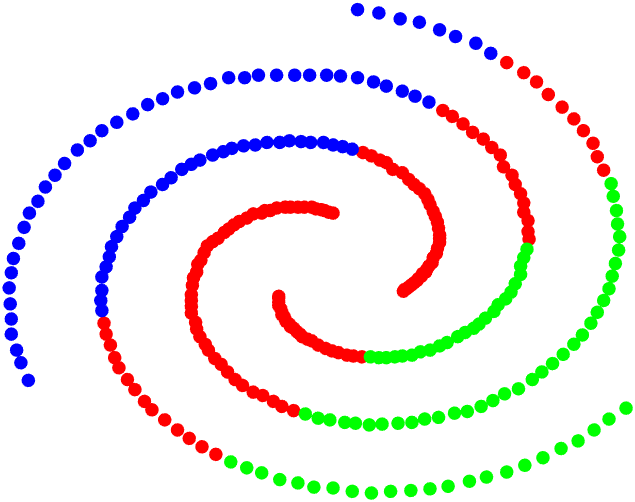}
\end{minipage}}
\quad
\subfigure[AWLRR]{\begin{minipage}[t]{0.102\linewidth}
\centering
\includegraphics[width=0.8in]{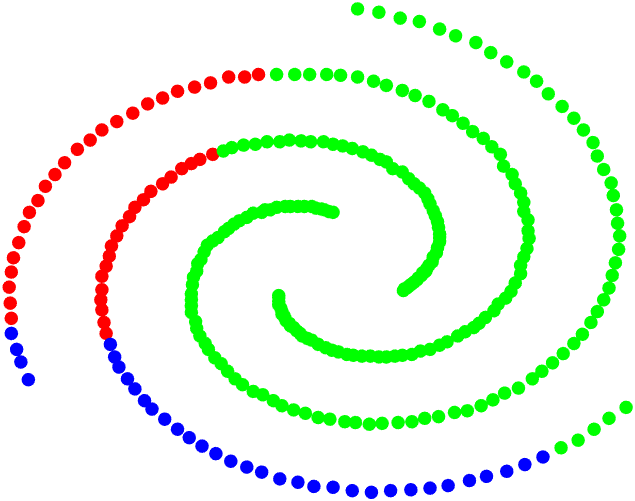}
\end{minipage}}
\quad
\subfigure[LRRAGR]{\begin{minipage}[t]{0.102\linewidth}
\centering
\includegraphics[width=0.8in]{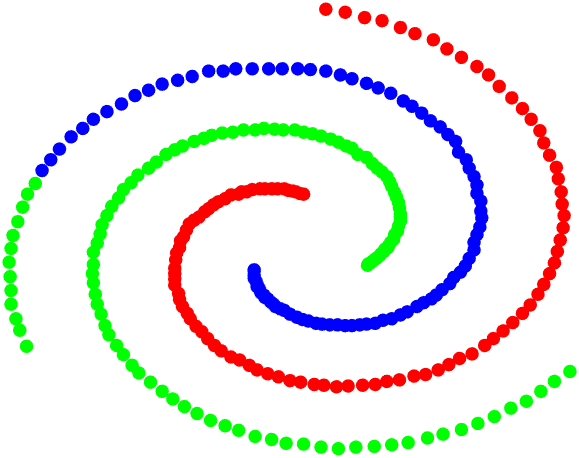}
\end{minipage}}
\quad
\subfigure[RSEC]{\begin{minipage}[t]{0.102\linewidth}
\centering
\includegraphics[width=0.8in]{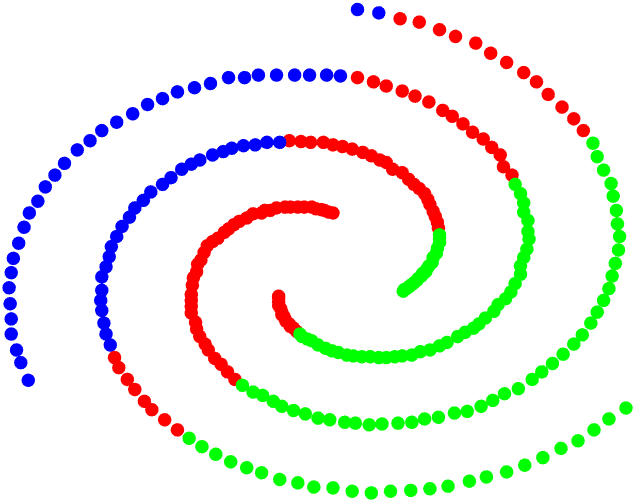}
\end{minipage}}
\quad
\subfigure[LapNR]{\begin{minipage}[t]{0.102\linewidth}
\centering
\includegraphics[width=0.8in]{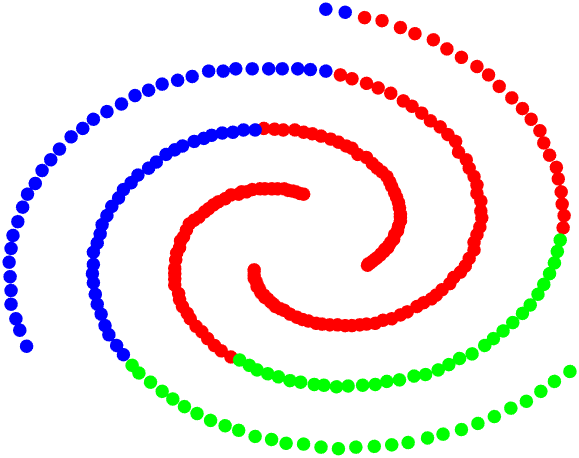}
\end{minipage}}
\quad
\subfigure[ALRR]{\begin{minipage}[t]{0.102\linewidth}
\centering
\includegraphics[width=0.8in]{spiral/truth.png}
\end{minipage}}
\caption{Experimental results on the spiral database.}
\label{spiraldatset}
\end{figure*}
\begin{figure*}[htbp]
\centering
\subfigure[LRR]{\begin{minipage}[t]{0.12\linewidth}
\centering
\includegraphics[width=0.8in]{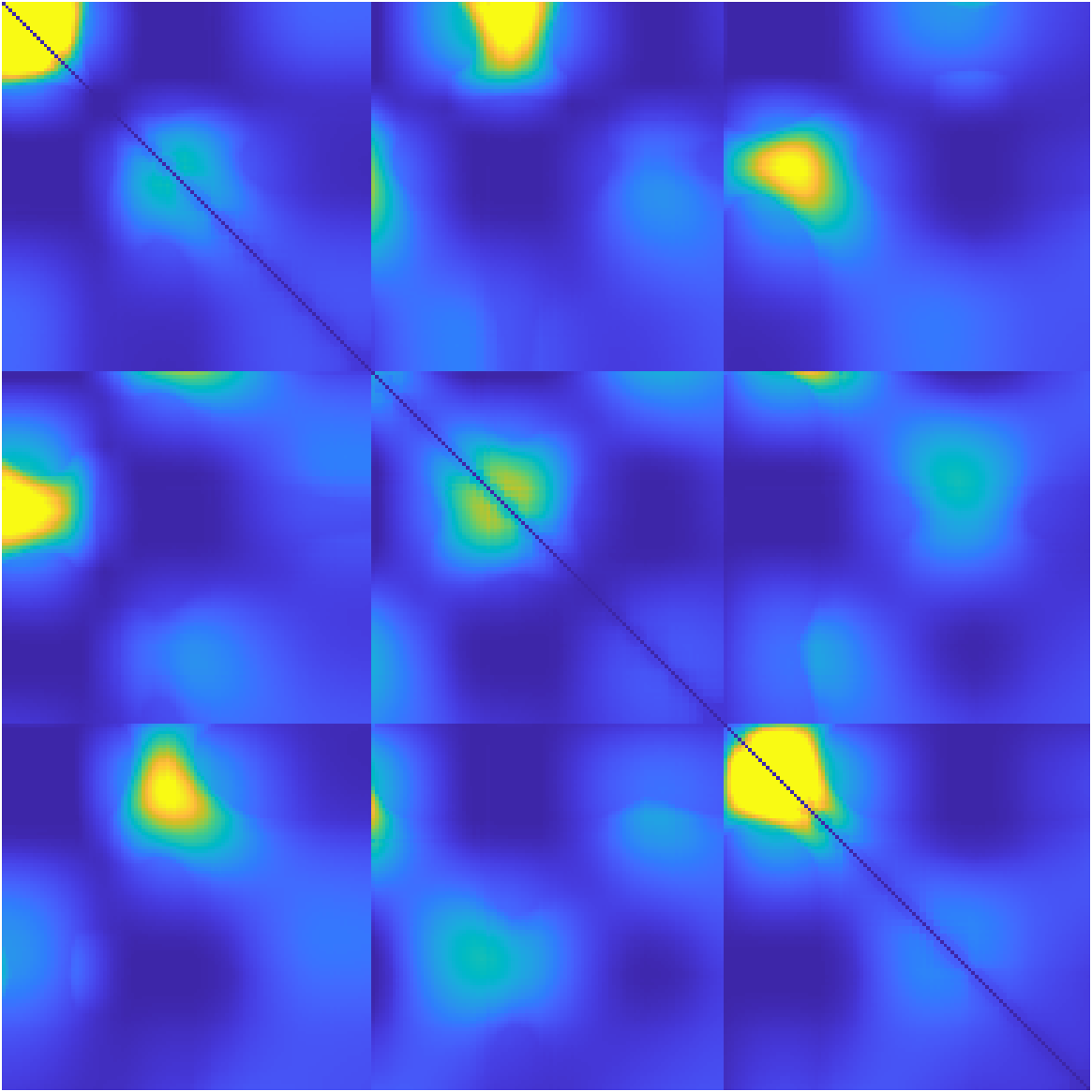}
\end{minipage}}
\quad
\subfigure[NSLLRR]{\begin{minipage}[t]{0.12\linewidth}
\centering
\includegraphics[width=0.8in]{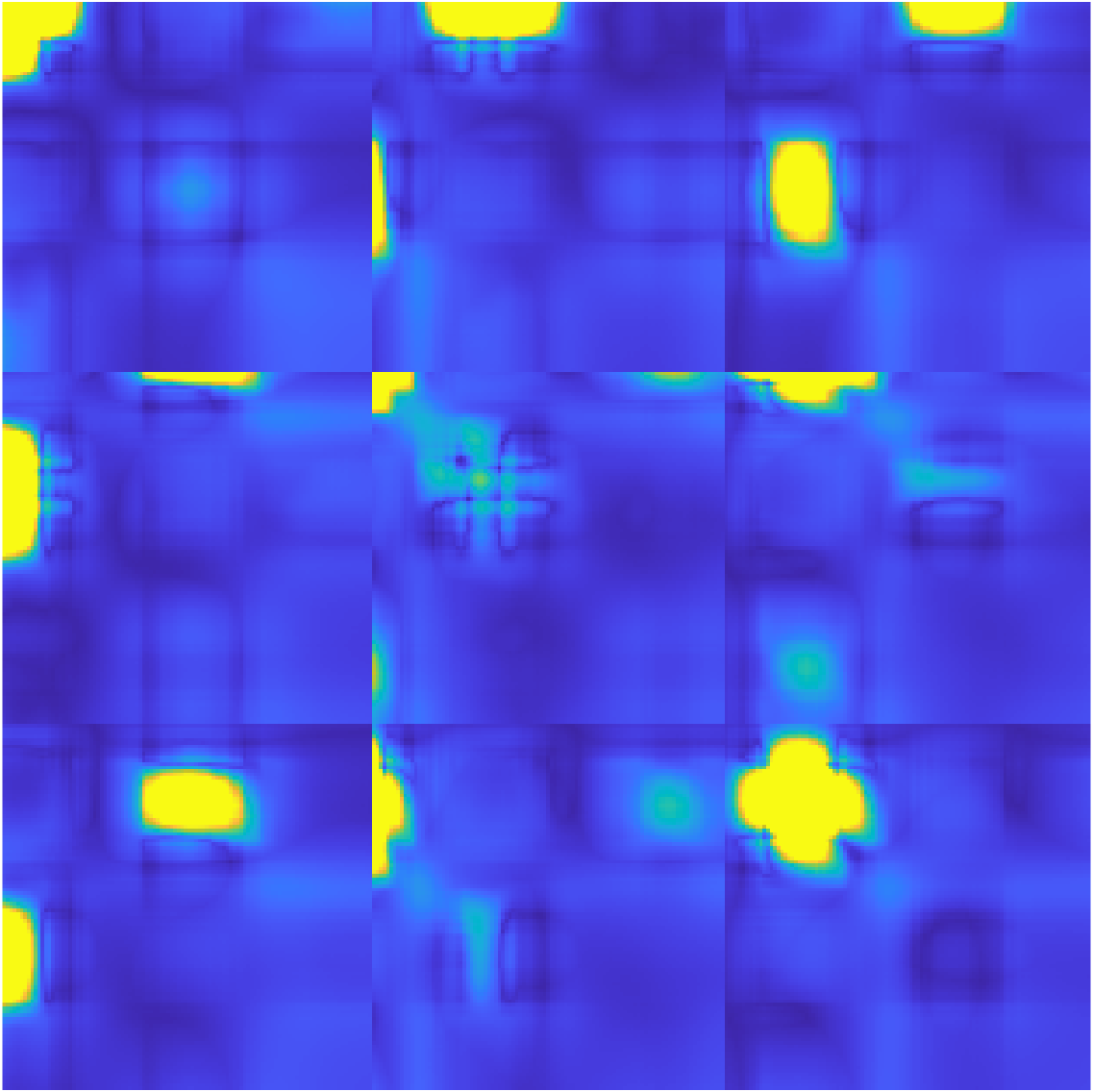}
\end{minipage}}
\quad
\subfigure[AWNLRR]{\begin{minipage}[t]{0.12\linewidth}
\centering
\includegraphics[width=0.8in]{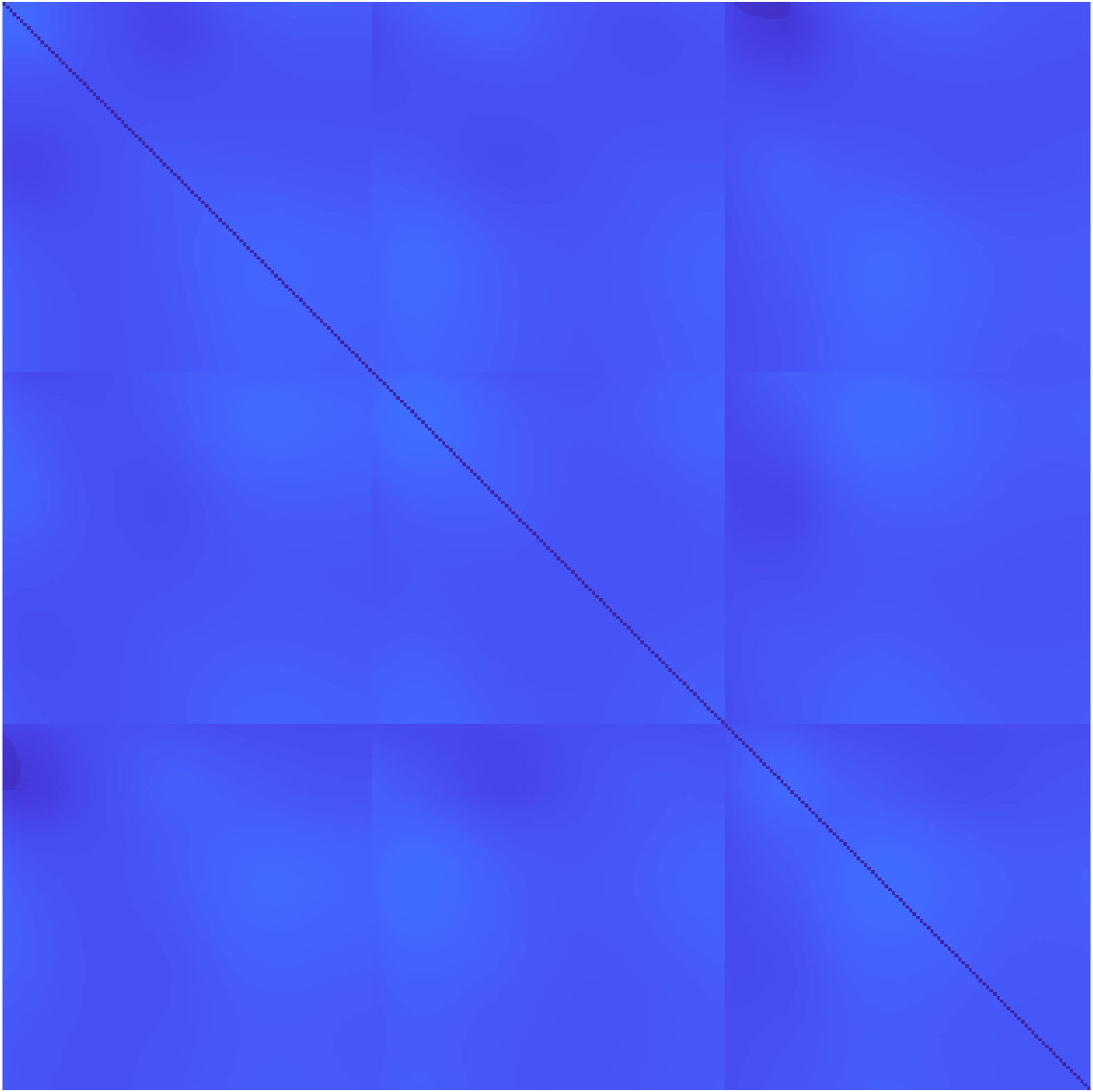}
\end{minipage}}
\quad
\subfigure[LRRAGR]{\begin{minipage}[t]{0.12\linewidth}
\centering
\includegraphics[width=0.8in]{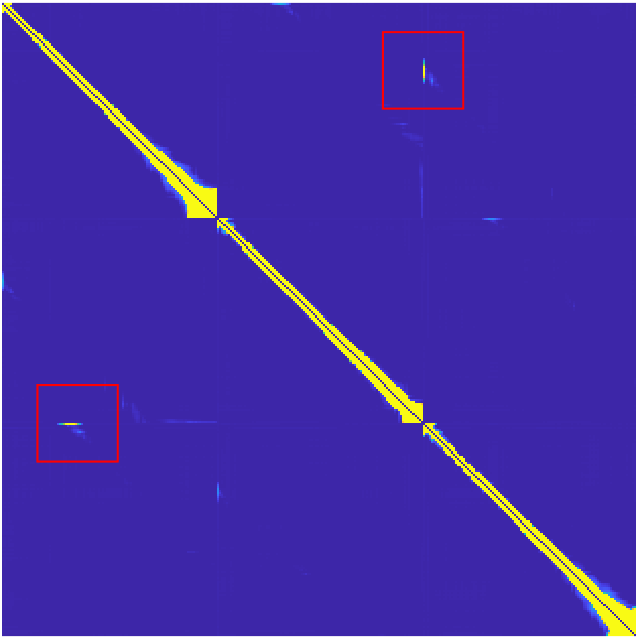}
\end{minipage}}
\quad
\subfigure[RSEC]{\begin{minipage}[t]{0.12\linewidth}
\centering
\includegraphics[width=0.8in]{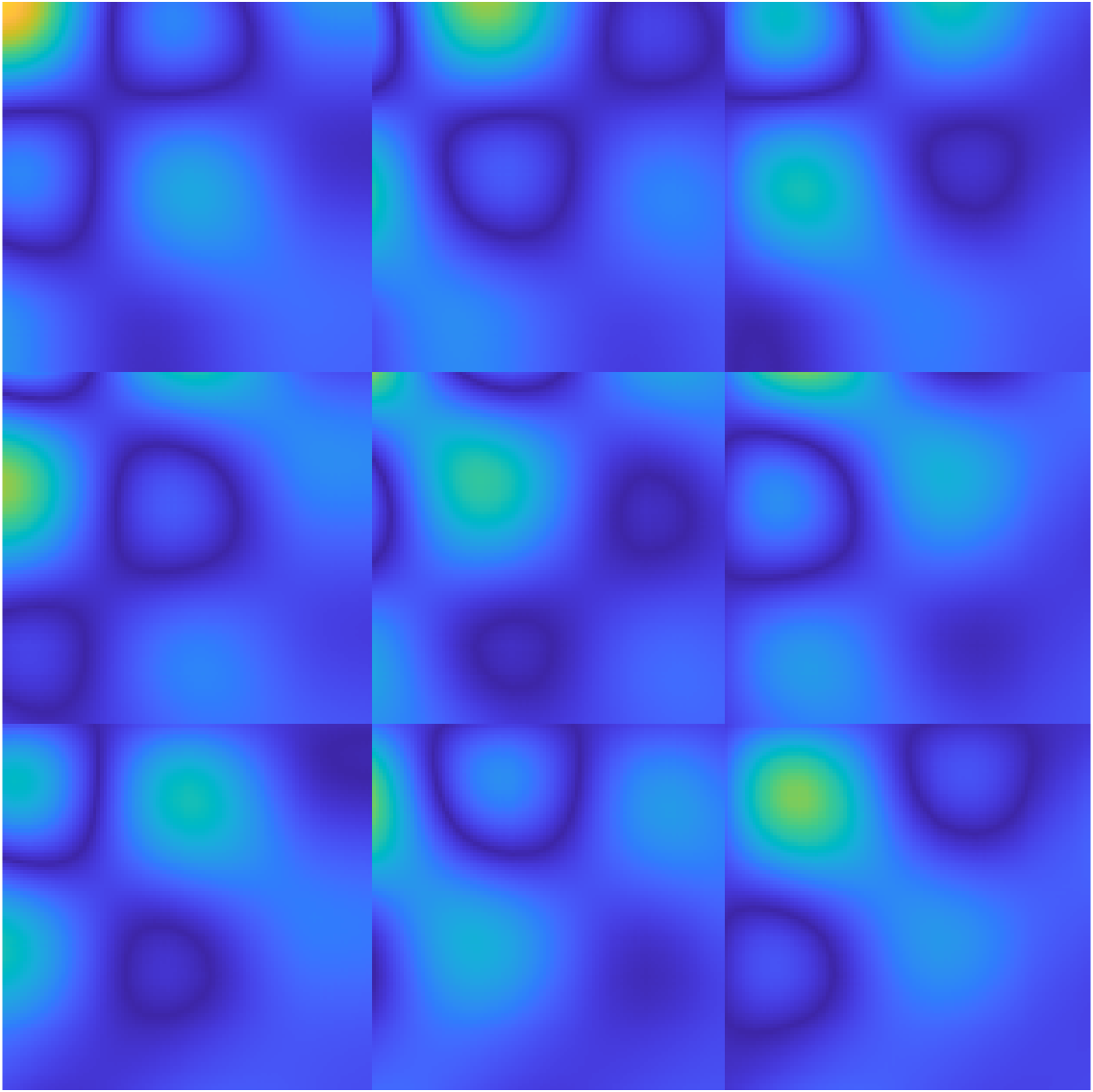}
\end{minipage}}
\quad
\subfigure[LapNR]{\begin{minipage}[t]{0.12\linewidth}
\centering
\includegraphics[width=0.8in]{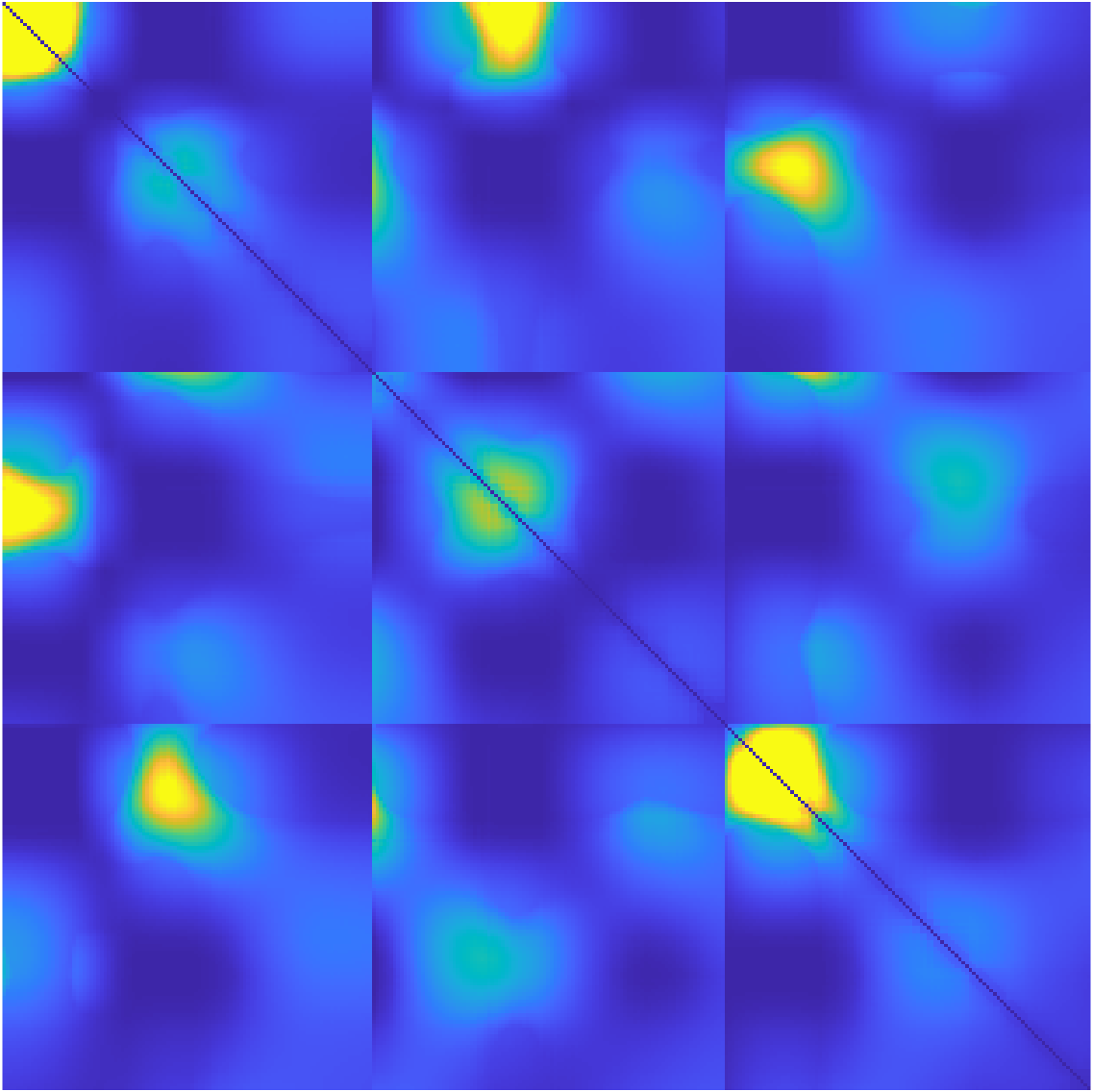}
\end{minipage}}
\quad
\subfigure[ALRR]{\begin{minipage}[t]{0.12\linewidth}
\centering
\includegraphics[width=0.8in]{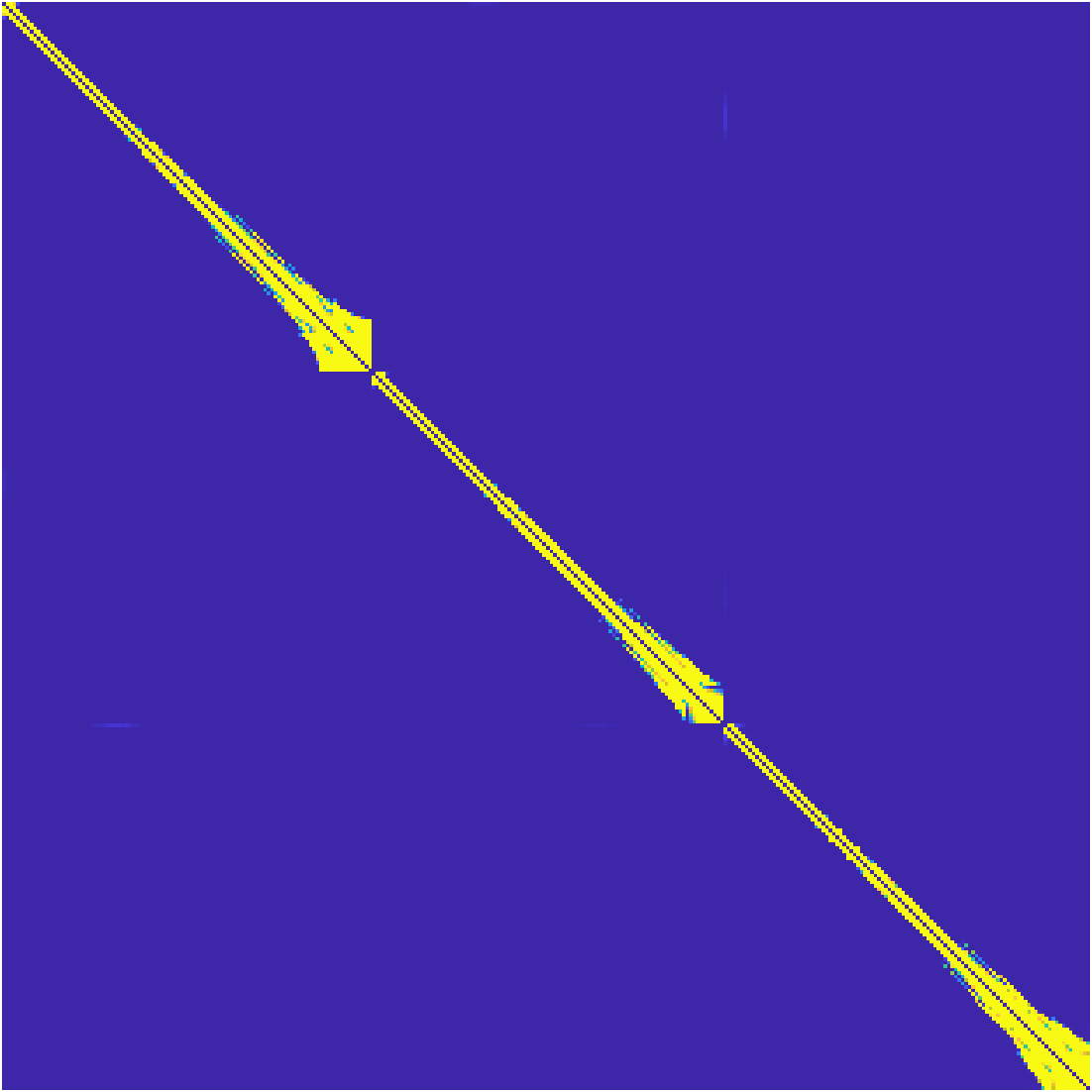}\label{ALRRr}
\end{minipage}}
\caption{Visual comparison of the SGs of LRR, NSLLRR, AWNLRR, RSEC, LapNR, HWLRR and our ALLR.}
\label{z}
\end{figure*}
In this subsection, the computational complexity is analysed firstly. As shown in Algorithm \ref{A1}, solving the proposed method contains six main steps, i.e., step \ref{step1} to step \ref{step6}. Here, the computational complexity of each step is analysed respectively. Step \ref{step1} is updated as Eq.(\ref{Ze}) in which $(X^TX+2I)^{-1}$ costs the most computational complexity, and its computational complexity is $\mathcal{O}(n^3)$. However, this term can be pre-calculated to reduce the computational complexity. Steps \ref{step2} and \ref{step3} use singular value thresholding (SVT) and eigen-decomposition respectively, thus the computational complexity of them are $\mathcal{O}(n^3)$ and $\mathcal{O}(cn^2)$, where $c$ is the number of learned rank. The computational complexity of step \ref{step4} and \ref{step5} is $\mathcal{O}(n^2)$. Since the computational complexities of basic matrix operations are much lower, these computational complexities are not taken into account. Finally, the computational complexity of the proposed method is $\mathcal{O}(\tau(n^3+(c+2)n^2))$, where $\tau$ is the number of iteration.
\vspace{-1em}
\begin{figure}[htbp]
\subfigure[Cars]{\begin{minipage}[t]{0.43\linewidth}
\centering
\includegraphics[width=1.6in]{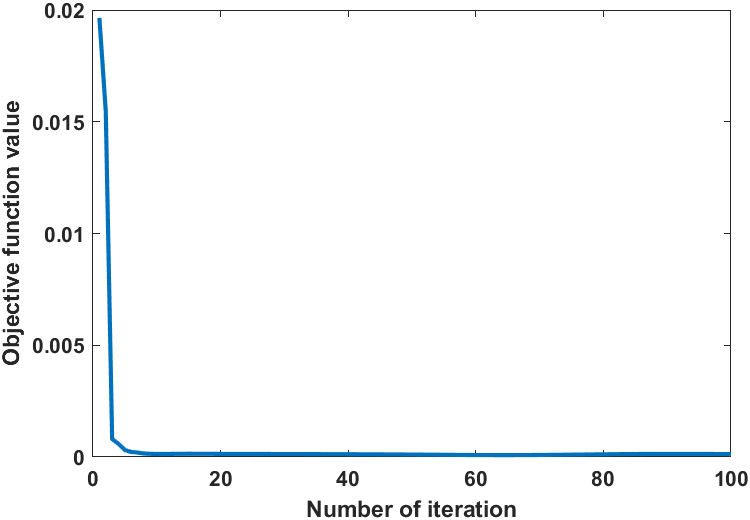}
\end{minipage}
}
\quad
\subfigure[Control]{\begin{minipage}[t]{0.43\linewidth}
\centering
\includegraphics[width=1.6in]{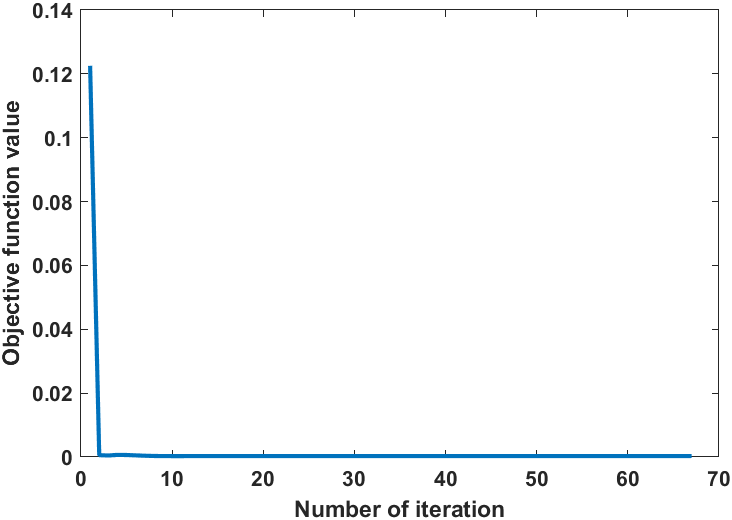}
\end{minipage}
}
\caption{Convergence curve of our ALRR on Cars and Control, in which all classes of each database are selected.}
\label{Convergence}
\end{figure}
\vspace{-1em}

The proposed method is solved by ADMM-style method with six blocks. The strongly convex of two-block ADMM method has been proved in \cite{LinLL15}. However, it is still hard to prove that a six-block ADMM method is convex. Hence, we prove the convergence of ALRR by experiments in the following. As shown in Fig.\ref{Convergence}, the value of the objective function, i.e., $\text{Obj}= (\sum_{i,j}\|Ax_i-Ax_j\|_2^2z_{i,j} +\lambda_1\|Z\|_*+\lambda_2\|E\|_1+\lambda_3\langle L_Z,Y\rangle)/\|X\|_F$, will monotonically decreasing and finally arriving the local optimum, which can show the convergence of ALRR.
\section{Experiments}
In this section, some experiments are conduceted to show the effectiveness of our ALRR method. Here, some self-representation methods, i.e., LRR, NSLLRR \cite{yin2015laplacian}, AWNLRR \cite{wen2018adaptive}, LRRAGR \cite{WenFXTF18}, RSEC \cite{tao2019robust} and LapNR \cite{LapNR} are chosen as the comparison algorithms. To make it fair, the parameters of each method are varied in a wide range to obtain the optimum performance. Moreover, the SGs obtained by these methods are symmetrized by $W=(|Z|+|Z|^T)/2$ and $W$ is handled by Ncut \cite{shi2000normalized} to obtain the clustering result. In the experiments, a synthetic database and some benchmark real databases are used to evaluate the performance of all the methods and the details of these databases are shown in Table \ref{database}.
\begin{table}[htbp]\small
\caption{Description of the databases}
\vspace{-1em}
\begin{center}
        \begin{tabular}{ccccc}
    \toprule
    Type  & Database & Samples & Dim   & Classes \\
    \midrule
    Synthetic &Spiral &393 &2 &3\\
    \midrule
    \multirow{5}[2]{*}{UCI}
          & Cars  & 392   & 8       & 3 \\
          & Contral & 600   & 60    & 6 \\
          & Isolet  &1560   &617    &2\\
          & Solar & 323      &12       &6   \\
          & Yeast & 1484  & 1470   & 10 \\
    \midrule
    \multirow{2}[2]{*}{Handwritten} & Dig   & 1797  & 64    & 10 \\
          & USPS  & 1000  & 256   & 10 \\
    \midrule
    \multirow{3}[2]{*}{Face} & Jaffe & 213   & 676   & 10 \\
    & Yale & 165   & 1024   & 15 \\
    & EYB\tnote{*} & 2414   & 1024   & 38 \\
    \bottomrule
    \end{tabular}
\end{center}  
\label{database}
  \begin{tablenotes}
       \tiny
       \item[*] EYB denotes Extended Yale B.
     \end{tablenotes}
\end{table}

\vspace{-1em}
\subsection{Clustering on synthetic database}
\begin{table*}[ht]
  \centering
  \caption{Clustering results on real databases}
        \resizebox{\textwidth}{!}{\begin{tabular}{lccccccccccc}
    \toprule
    Database & Metric & Cars  & Control & Isolet & Solar & Yeast & Dig   & USPS  & Jaffe & Yale  & EYB \\\midrule
    Ncut  & \multirow{8}[1]{*}{ACC} & 48.72 & 51.50 & 55.58 & 51.7  & 32.28 & 76.85 & 49.50 & 89.20 & 20.00 & 19.72 \\
    LRR   &       & 62.76 & 48.17 & 55.96 & 51.39 & 30.26 & 79.13 & 53.30 & 99.53 & 46.06 & 67.44 \\
    NSLLRR &       & 63.52 & 65.00 & 59.36 & 54.80 & 39.22  & 67.78 & 54.20 & 99.53  & 54.55 &38.48  \\
    AWNLRR &       & 66.33 & 53.17 & 58.40 & 55.11 & 10.71 & 79.86 & 55.00 & 98.59 & 41.84 & 88.07 \\
    LRRAGR &       & 62.76 & 56.83 & 54.00 & 45.51 & 30.39 & 59.32 & 40.90 & 98.59  & 56.36 &87.04  \\
    RSEC  &       & 63.01 & 54.33 & \textbf{62.95} & 56.04 & 38.01 & 79.19 & 53.80 & \textbf{100}   & 55.15 & 88.53 \\
    LapNR &       & 57.14 & 37.83 & 58.65 & 52.63 & 39.22  & 76.02 & 57.20 & 98.12 & 55.15 & 48.76 \\
    ALRR  &       & \textbf{68.11} & \textbf{74.50} &61.47 & \textbf{59.75} & \textbf{44.54} & \textbf{82.80} & \textbf{59.40} & \textbf{100} & \textbf{61.21} & \textbf{99.13} \\
    \midrule
    Ncut  & \multirow{8}[2]{*}{Fscore} & 48.60 & 58.35 & 50.62 & 43.03 & 38.47 & 67.71 & 36.96 & 82.45 & 14.43 & 12.79 \\
    LRR   &       & 63.17 & 57.25 & 50.65 & 44.82 & 35.93 & 72.85 & 43.82 & 99.05 & 28.25 & 46.18 \\
    NSLLRR &       & 63.67 & 62.11 & 51.75 & 45.87 & 29.39 & 63.23 & 42.35 & 99.03 & 36.79 & 13.12 \\
    AWNLRR &       & 66.04 & 53.47 & 51.53 & 46.65 & 28.96 & 76.90 & 46.81 & 97.11 & 38.57 & 85.14 \\
    LRRAGR &       & 59.25 & 69.10 & 51.00 & 37.84 & 35.94 & 45.25 & 38.90 & 97.10 & 38.68 &78.63  \\
    RSEC  &       & 58.92 & 57.61 & 53.43 & 47.81 & 31.75 & 72.32 & 44.33 & \textbf{100}   & 35.07 & 80.82 \\
    LapNR &       & 50.54 & 54.28 & 52.24 & 45.62 & 29.39  & 71.70 & 47.24 & 96.32 & 35.40 & 25.95 \\
    ALRR  &       & \textbf{67.47} & \textbf{75.17} & \textbf{53.98} & \textbf{50.65} & \textbf{32.84} & \textbf{75.19} & \textbf{48.39} & \textbf{100} & \textbf{44.76} & \textbf{98.26} \\
    \bottomrule
    \end{tabular}}
  \label{result}%
\end{table*}%

A spiral database \cite{ChangY08} shown in Fig.\ref{ori} is used to evaluate the performance of ALRR and the comparison methods. This database contains three clusters, and many samples with different label are close in this synthetic database. Thus, using this database can show the ability of clustering methods handling the nearby samples with different labels. As shown in Fig.\ref{ALRRr}, ALRR can correctly divide the samples into three clusters against the misleading of the nearby samples with different labels, and the other methods have assigned wrong labels to some sample. To further show the discrimination of the learned SGs, the SGs obtained by all the self-representation is shown by visualization. Here, all the SGs obtained have been symmetrized by $W=(|Z|+|Z|^T)/2$. From Fig.\ref{z}, we can find that LRRAGR can learn a SG with three parts, which performs much better than the other comparison methods. However, the SG learned by LRRAGR contains some similarities among samples with different labels are greater than 0, which can mislead the clustering method and leads to a worse performance. Due to the learned sparse SG with exactly three diagonal blocks, ALRR can achieve the best clustering result.

\subsection{Clustering on real databases}
In this subsection, some benchmark real databases are used to evaluate the performance of the proposed method and comparision methods. Two most used metrices, i.e., ACC and Fscore, are utilized to compare the performance of the final clustering results. 

The experimental results on these real databases are given in Table \ref{result}, and we can find some conclusions as follows.

\begin{itemize}
\setlength{\itemsep}{0pt}
\setlength{\parsep}{0pt}
\setlength{\parskip}{0pt}
    \item Overall, the proposed ALRR outperforms the comparison methods on most databases and can obtain competitive results on the other databases, which can prove the effectiveness of ALRR. Specifically, for the databases with more dimensions, e.g., EYB and Yale, ALRR performs much better than the other methods which prove that the proposed method is more effective on the high-dimensional database. This is because high-dimensional data contains more redundant features, and the auto-weighted matrix can enlarge the effect of the discriminative features.
    \item Compared with LRR, NSLLRR, LRRAGR, AWNLRR, HWLRR and ALRR perform better in the most cases. Since LRR just uses a global low-rank constraint to capture the global, NSLLRR, AWNLRR, HWLRR and ALRR improve the LRR by preserving more local structure. Thus, it is obvious that learning the local structure is effective for clustering task. 
    \item From the comparison among AWNLRR, LRRAGR and ALRR, we can find that ALRR obtains higher accuracy. These three methods use the distance penalty to learn more geometric structure, but ALRR uses an auto-weighted penalty to enlarge the effect of the disciminative features, which leads to a better SG.
    \item LRRAGR and ALRR both take use of the class information. LRRAGR utilizes the class information by a rank constraint, and ALRR ensures that the learned SG contains $k$ diagonal blocks. Hence, this can prove that the block constraint is more effective than the rank constraint for clustering.
\end{itemize}
From these analyses, the effectiveness of the auto-weighted penalty and the block constraint have been proved. With the integration of above factors, the proposed ALRR performs better than the other methods.

\subsection{Effectiveness of the auto-weighted matrix}
\begin{figure*}[ht]
\centering
\subfigure[Cars]{\begin{minipage}[t]{0.3\linewidth}
\centering
\includegraphics[width=2in]{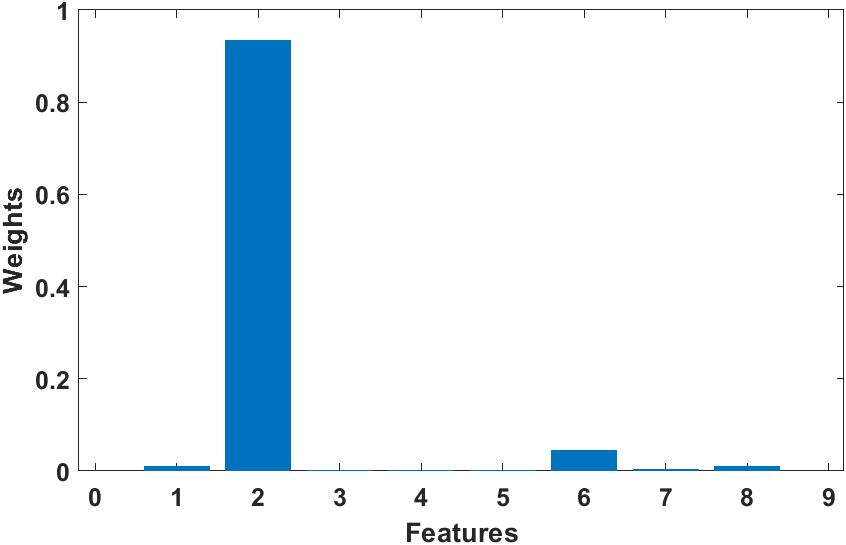}
\end{minipage}}
\quad
\subfigure[Control]{\begin{minipage}[t]{0.3\linewidth}
\centering
\includegraphics[width=2in]{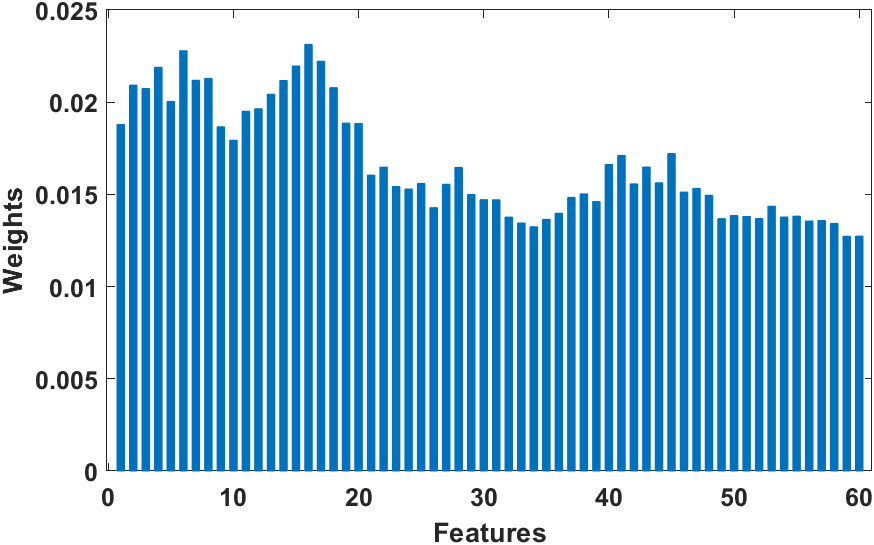}
\end{minipage}}
\quad
\subfigure[Ecoli]{\begin{minipage}[t]{0.3\linewidth}
\centering
\includegraphics[width=2in]{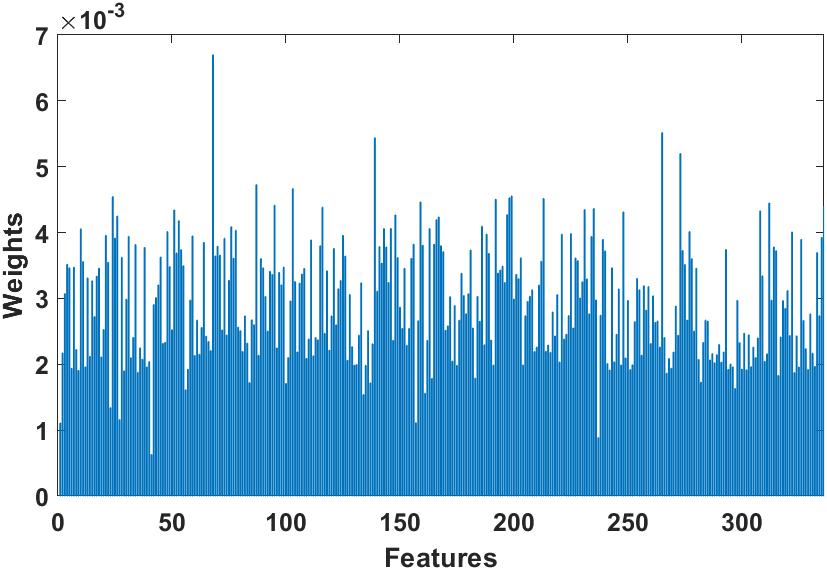}
\end{minipage}}
\quad
\subfigure[Solar]{\begin{minipage}[t]{0.3\linewidth}
\centering
\includegraphics[width=2in]{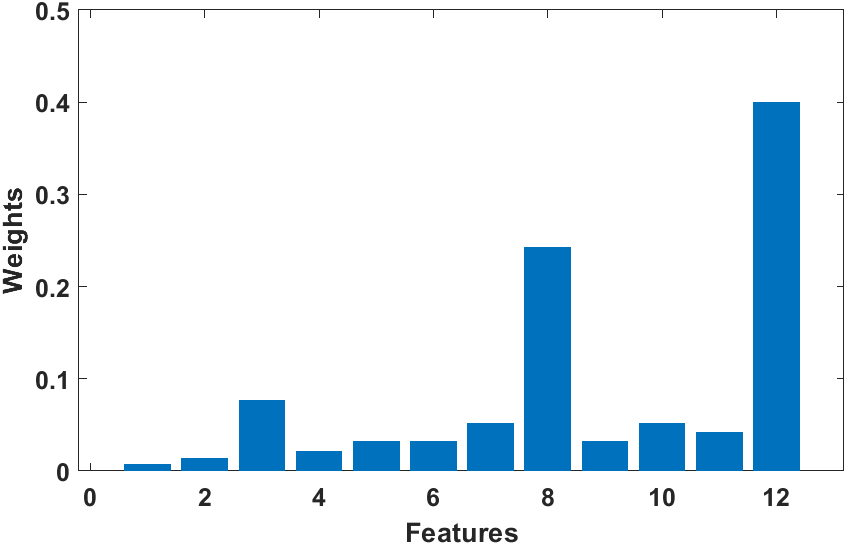}
\end{minipage}}
\quad
\subfigure[Yeast]{\begin{minipage}[t]{0.3\linewidth}
\centering
\includegraphics[width=2in]{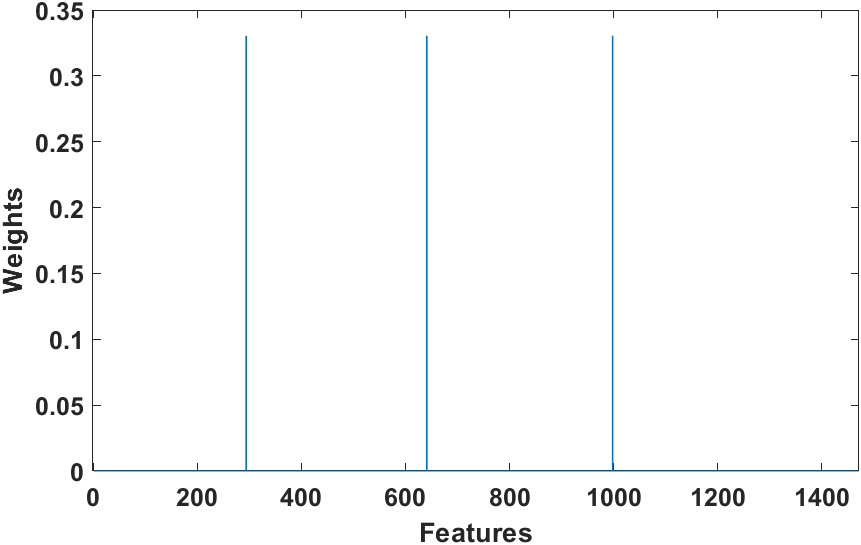}
\end{minipage}}
\quad
\subfigure[PD]{\begin{minipage}[t]{0.3\linewidth}
\centering
\includegraphics[width=2in]{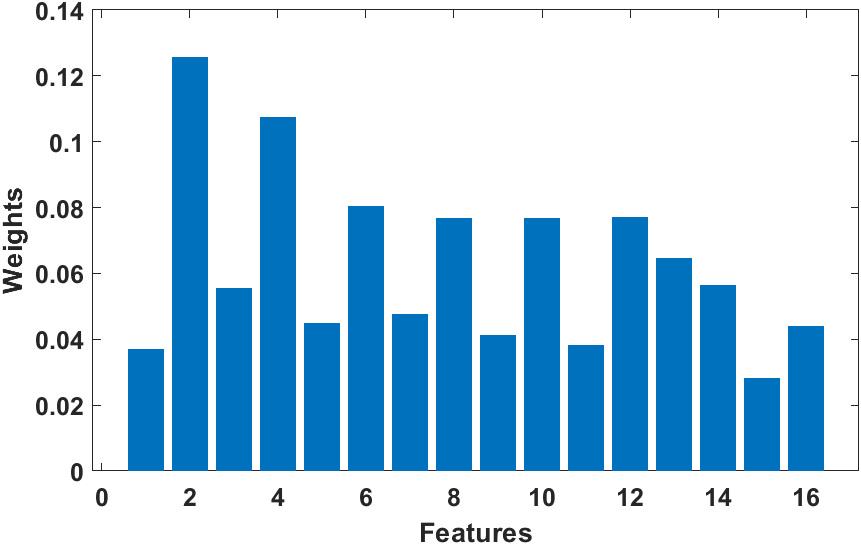}
\end{minipage}}
\caption{The learned weights of the databases.}\label{auto}
\end{figure*}

\begin{table*}[htbp]
  \centering
  \caption{Clustering results on original and weighted features}
    \begin{tabular}{ccccccccccc}
    \toprule
     & Cars & Control & Isolet & Solar & Yeast & Dig & USPS & Jaffe & Yale & EYB \\
    \midrule
    Original&66.82 &59.00 &59.32 &56.54 &41.58 &78.50 &53.80 &98.59 &57.58 &80.36 \\
    Weighted&68.11 &74.50 &61.47 &59.75 &44.54 &82.80 &59.40 &100 &61.21 &99.13 \\
    \bottomrule
    \end{tabular}%
  \label{autowei}%
\end{table*}%
To further show the effectiveness of the auto-weighted matrix, some learned auto-weighted matrix are shown in Fig.\ref{auto}. It can seen that the weights of different features are different, and the weights are adaptively assigned as 1) if the database just contains a few discriminitive features (e.g., Cars and Yeast), the auto-weighted matrix will just select the most important features and remove the useless features; 2) for the database with all the features useful (e.g., Control and PD), the auto-weighted matrix can assign more reasonable weights to enhance the features. Furthermore, we show the contribution of the auto-weighted matrix in our method by setting the auto-weighted matrix $A=I$ in ALRR. As shown in Table \ref{autowei}, the clustering results on the weighted features are better than that on the original features, which shows the effectiveness of the auto-weighted penalty.
\subsection{Parameter sensitivity and selection}
\begin{figure}[htbp]
\subfigure[$\lambda_1$]{\begin{minipage}[t]{0.43\linewidth}
\centering
\includegraphics[width=1.5in]{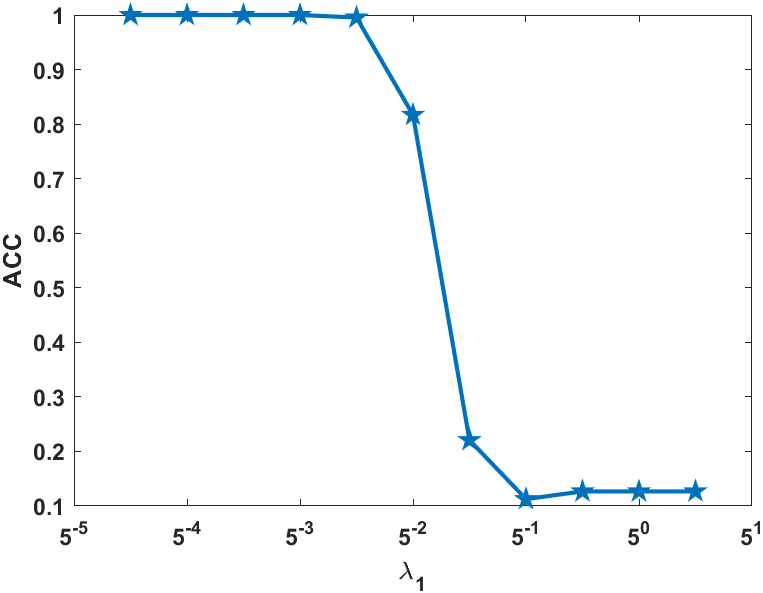}\label{1}
\end{minipage}}
\quad
\subfigure[$\lambda_2$ and $\lambda_3$]{\begin{minipage}[t]{0.43\linewidth}
\centering
\includegraphics[width=1.5in]{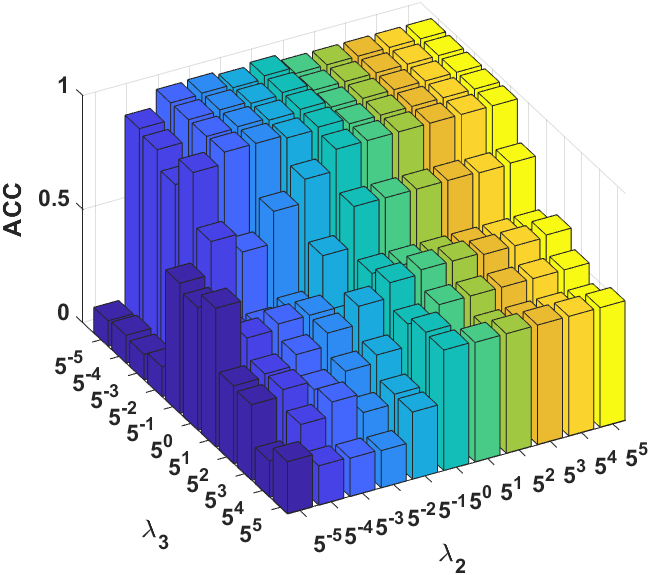}\label{23}
\end{minipage}}
\caption{Parameter sensitivity analysis of ALRR on the Jaffe,
where (a) fix $\lambda_2$ and $\lambda_3$ to tune $\lambda_1$; (b) fix $\lambda_1$ to tune $\lambda_2$ and $\lambda_3$.}
\end{figure}
\vspace{-1em}
As shown in model (\ref{final}), there are three parameters, i.e., $\lambda_1$, $\lambda_2$ and $\lambda_3$ in the ALRR. They are used to balance the effect of low-rank constraint, error and block constraint, respectively. In this section, the sensitivity of each parameter is tested by performing the proposed method with different combinations of three parameters, and each parameter is varied in a wide range $[5^{-5},5^{-4},...,5^4,5^5]$. First, we fix $\lambda_2=5^{-2}$ and $\lambda_3 = 5^{-2}$ to tune $\lambda_1$, and thus the sensitivity of $\lambda_1$ is shown as Fig.\ref{1}. It is obvious that ALRR can deliver good results with $\lambda_1\leq 5^{-1}$. Then, $\lambda_1$ is fixed as $5^{-2}$, and the influence of $\lambda_2$ and $\lambda_3$ is showed by performing the proposed method with different combinations of $\lambda_2$ and $\lambda_3$ on the Jaffe database. As shown in Fig.\ref{23}, we can find that ALRR performs well with $\lambda_2\leq 5^{-2}$ and $\lambda_3 \leq 5^{-2}$. Since finding a suitable combination of parameters is still an open problem, and we just confirm that the most suitable parameters in our method can be found in a small range, i.e., $[5^{-5},5^{-4},5^{-3},5^{-2}]$.
\section{Conclusion}
In this paper, a novel and unsupervised self-representation learning method, i.e., Auto-weighted Low-Rank Representation (ALRR), is proposed. Our ALRR can learn a discriminative SG which contains $k$ diagonal blocks which is a clear clustering structure. With the guidness of this term, the auto-weighted penalty can adaptively assign different weights to the features which can enlarge the effect of the useful features and reduce the impact of the useless features. Moreover, this penalty can preserve more local structure with the weighted features. The effectiveness of our ALRR for clustering has been examined on both synthetic and real databases.

\bibliographystyle{named}
\bibliography{ijcai21}
\end{document}